\documentclass{article} 
\usepackage[preprint]{neurips_2025}
\usepackage[utf8]{inputenc}
\usepackage[T1]{fontenc}
\usepackage[colorlinks=true, linkcolor=blue, citecolor=blue, urlcolor=blue]{hyperref}
\usepackage{url}
\usepackage{booktabs}
\usepackage{amsmath}
\usepackage{amsfonts}
\usepackage{nicefrac}
\usepackage{microtype}
\usepackage{xcolor}
\usepackage{graphicx}
\usepackage{makecell}
\usepackage{listings}
\usepackage{amsmath,amssymb}
\usepackage{algorithm}
\usepackage{algpseudocode}
\usepackage{bm}

\usepackage{caption}   
\usepackage[most]{tcolorbox}

\graphicspath{{figures/}}

\usepackage[most]{tcolorbox}
\usepackage{tikz}
\usepackage{xcolor}

\definecolor{contribgreen}{RGB}{91,137,78}
\definecolor{contribbg}{RGB}{242,248,237}

\newtcolorbox{Prompts}{
  enhanced,
  breakable,
  colback=contribbg,
  colframe=contribgreen,
  boxrule=0pt,
  leftrule=3.5pt,
  arc=0pt,
  outer arc=0pt,
  left=8pt,
  right=8pt,
  top=7pt,
  bottom=7pt,
  boxsep=0pt,
  before skip=8pt,
  after skip=8pt
}

\title{Revisiting the ``Push-T'' \\ Robot Manipulation Task with Agentic Robotics}

\author{
Shuangyu Xie,~~~ Kaiyuan Chen,~~~ Ken Goldberg\\[4pt]
University of California, Berkeley \\[2pt]
\texttt{\{syxie, kych, goldberg\}@berkeley.edu}
}
\begin{document}
\maketitle


\begin{abstract}
Push-T is an iconic benchmark for learning   manipulation policies from human demonstrations. 
The robot must use a single point of contact to push a T-shaped block into a target pose. 
In this short paper, we revisit the Push-T task in the context of emerging advances in Agentic Robotics where an LLM coding agent  -- Claude Code with Fable 5  -- is prompted to create an algorithmic solution that does not require any demonstration data. 
We study how effective the agentic coding loop can solve the Push-T task, and compare the resulting code as policy with the visuomotor imitation learning policy. 
Results suggest that the agent found the 2D gym simulation online, and used sim experiments to learn push mechanics, iteratively optimizing to achieve 100\% success rate using 50\% fewer steps than the best diffusion policy trained with 200 human demonstrations. 
The coding agent also solve extensions from T to the full alphabet (Push-A to Push-Z) using a self generated curriculum and generated simulation code for the Franka and UR5 robot arms in 3D cross-embodiment simulations with visual feedback. 
Videos, policies and details will be posted online.

\end{abstract}

\section{Introduction}
\label{sec:intro}

Push-T is an iconic planar task in robot learning. Introduced by the Implicit Behavioral Cloning paper in 2021 ~\cite{florence2021implicit} and popularized by the Diffusion Policy paper \cite{chi2023diffusion} in 2023, a robot must use a single point of contact to push a T-shaped block into a target pose. Because Push-T is planar, multimodal, and easy to simulate and evaluate, it has primarily been utilized to explore imitation learning with diffusion models. 



Beyond visuomotor policy learning, classical work established geometric foundations for designing contact actions that orient polygonal objects~\cite{goldberg1993orienting}. Later work showed that, under a fixed pushing contact, planar pushing can be
reduced to a Dubins-car system, yielding bounded-curvature, time-optimal
pushing trajectories~\cite{Pushing}. ~Extending such procedural methods to Push-T requires reasoning about its concave geometry, frictional contacts, and coupled translation and rotation. CRISP~\cite{crisp2025} addresses these challenges by formulating Push-T as a contact-implicit trajectory-optimization problem solved through sequential convex programming. Interactive World Simulator~\cite{interactiveworldsim2026} takes a data-driven approach, learning an action-conditioned world model of Push-T interactions for policy training and evaluation. Despite their different formulations, these methods require substantial task-specific effort in algorithm design, optimization modeling, or interaction-data collection.


We revisit the Push-T task in the context of \emph{agentic robotics}, where recent advances in large language models (LLMs) generate robot commands without requiring the LLM to operate directly in the high frequency control loop \cite{liang2023code}~\cite{fu2026capx}  \cite{chen2026gap}.
Rather than learning a task-specific policy from demonstrations or relying on a manually derived contact model, an agentic system can iteratively construct, execute, evaluate, and refine a procedural controller. We ask: Can AI coding agent generate procedural methods to solve the  Push-T task and if so, how do these solutions compare with policies learned from human demonstrations? As a further step, we then extend the problem to Push-Alphabet: can the agent systematically adapt its geometric reasoning, contact strategies, and control logic to solve a diverse set of concave and asymmetric character shapes at scale?

\section{Problem Setup}
We study planar pushing with a single point-contact pusher. Let
$x \in \mathcal{X}$ denote a rigid object, where
$\mathcal{X}=\{\mathrm{T}\}$ for \emph{Push-T} and
$\mathcal{X}=\{\mathrm{A},\ldots,\mathrm{Z}\}$ for
\emph{Push-Alphabet}. At time $t$, the environment has state
$s_t \in \mathcal{S}$, including the object pose, pusher configuration,
and target pose. A controller $\pi$ receives an observation
$o_t \in \mathcal{O}$ and generates an action
$a_t \in \mathcal{A}$ that approaches, contacts, and pushes the object.
The objective is to move the object into its target pose.

\noindent \textbf{Problem Variants.} We evaluate the task along two orthogonal axes: object diversity and
observation modality. The object is either the canonical T-shaped block
or one of 26 letter-shaped blocks. The controller receives either
\emph{privileged state}, containing the object, pusher, and target poses,
or a visual observation $I_t$, from which these quantities must be
estimated. This yields four settings:
\emph{Push-T + state}, \emph{Push-T + vision},
\emph{Push-Alphabet + state}, and \emph{Push-Alphabet+ vision}.

\noindent \textbf{Agentic Self-Learning.}
We task an LLM coding agent with synthesizing a procedural controller
$\pi_{\phi}$, where $\phi$ denotes the generated perception, planning,
dynamics, and control code. The agent iteratively executes the controller
in simulation, observes successes and failures, diagnoses model or control
errors, and revises $\phi$. We refer to this cycle of code generation,
execution, evaluation, and revision as \emph{agentic self-learning}.
The resulting controller may use analytical geometry, simulation-based
rollouts, system identification, and receding-horizon replanning, but the
LLM does not operate inside the high-frequency control loop.

\noindent \textbf{Metrics.} We evaluate each generated controller using two separate metrics: \emph{success rate} and \emph{push efficiency}. An episode is successful if the final object pose satisfies the task-specific position and orientation tolerances with respect to the target pose. Push efficiency is measured by the number of discrete pushes required in successful episodes.










\section{Solving Push-T with Agentic Robotics}
\label{sec:selfevo}

\subsection{Initial Solution}
To solve the  Push-T task, we use an autonomous agentic coding loop driven by Claude Code with Fable 5. Given a basic task description using natural language as follow, the agent search for the gym simulation environment for Push-T, iteratively generates procedural code, observes execution outcomes, and refines its logic. Crucially, the agent does not blindly assume idealized contacts; it actively runs candidate parameters through the simulation to fit the friction behavior and construct an internal quasi-statics model.

\begin{Prompts}
    \textbf{Task Description:} I want you to use code as policy to achieve Push-T. You are not allowed to use a learned policy, but using existing robot data is fine.
\end{Prompts}

Given the prompt, the controller generated by Claude Code uses a scripted state machine with four phases: plan, approach, push, and retreat. The code represents the environment state as a five-dimensional vector:
$o_t = \left[p^a_x,\; p^a_y,\; p^T_x,\; p^T_y,\; \theta^T\right],$
where $(p^a_x,p^a_y)$ is the pusher position and $(p^T_x,p^T_y,\theta^T)$ is the T-block pose.

Specifically, Claude Code creates a geometry-based hybrid feedback controller ( \texttt{controller\_v0} ) that alternates between contact selection, collision-free approach, bounded pushing, and retreat. It exploits a structural decomposition of planar pushing: face-normal pushes through the center of mass primarily translate the object, whereas offset pushes generate rotation through their moment arms. A discrete contact library provides both translational actions and clockwise or counterclockwise rotational actions. At each replanning step, the controller selects the dominant pose-error component, filters contacts by geometric reachability and boundary constraints, and greedily chooses the action predicted to reduce that error under a quasi-static contact model. Short-horizon execution followed by state-feedback replanning corrects model mismatch, while recovery pushes move the object away from configurations where useful faces are blocked by the arena boundary.

\subsection{Closed-Loop Perception Pipeline for Code-as-Policy}
\label{sec:perception}
After we identify the code doesnot use vision feedback in  \texttt{Claude\_state\_controller}, 
to eliminate reliance on privileged ground-truth geometric states, we ask the coding agent to add vision feedback into the perception loop:
\begin{Prompts}
    \textbf{Task Description:} Add vision feedback to PushT controller instead of directly taking the geometrical states. 
\end{Prompts}

Claude code end up with rewrite the controller to connect with the perception stack, which processes raw RGB input frames through a structured template fitting three-phase pipeline:

\begin{itemize}
    \item \textbf{Per-Frame Workspace Perception:} For every frame, the camera feed undergoes color segmentation against a pre-defined render palette. A physical sensor model erodes the segmented block mask and excuses the pusher body to isolate the block, goal, and agent masks. Closed-loop tracking then updates object poses via coarse-to-fine Intersection-over-Union (IoU) fitting.
    
    \item \textbf{One-Time Calibration:} Before manipulation starts, the agent fits a free scale parameter to match letter geometry, resolves multi-basin goal poses under potential occlusion, and initializes the parameters (polygon bounds, center of mass, inertia, and contact points).
    
    \item \textbf{Per-Push Parameter Adaptation:} Following each completed pushing segment, the controller buffers state setpoints, replays predicted orientation changes ($\Delta\theta$) in a digital twin, and refines the friction gain parameters: $f \leftarrow f \cdot r^{-\eta}$ where $f$ represents the friction gain factor updated with a decreasing rate driven by $r$ and $\eta$.
\end{itemize}
We denote the controller created by Claude Code using vision as \texttt{Claude\_vision\_controller}.

\subsection{Evaluation}
\paragraph{Baselines}
We compare a learned diffusion policy with procedural controllers created by Claude Code. The diffusion-policy baseline~\cite{chi2023diffusion} is trained specifically for \texttt{gym\_pusht} using 200 human demonstrations. We evaluate the publicly
released LeRobot Diffusion Policy checkpoint rather than retraining the model. It processes $96{\times}96$ image observations and the pusher position using a ResNet-18 visual encoder and a U-Net denoiser. 

\paragraph{Results}
Table~\ref{tab:pusht_results} reports performance on
\texttt{gym\_pusht} over 200 fixed evaluation seeds. The evaluated LeRobot
checkpoint achieves a 62.5\% success rate under the environment's stringent
criterion of at least 95\% target coverage, which is comparable to the 65.4\%
success rate reported by the checkpoint's official evaluation over 500
episodes. The mean final target coverage further explains this difference:
Diffusion Policy reaches 87.4\% coverage on average, often placing the object
close to the target without exceeding the 95\% success threshold, whereas the
generated code policy reaches 96.8\% mean coverage and therefore satisfies the
success criterion more consistently.
The code policy also requires substantially fewer interactions, reducing the mean episode length from 223.7 to 120.1 control steps and the mean number of pushes from 6.38 to 3.72. These results suggest that explicit geometric contact reasoning combined with closed-loop replanning can provide an effective alternative to demonstration-based visuomotor policy learning.

\begin{table}[htbp]
  \caption{
  Performance on \texttt{gym\_pusht} over 200 seeds.
  }
  \label{tab:pusht_results}
  \centering
  \small
  \setlength{\tabcolsep}{6 pt}
  \begin{tabular}{@{}lcccc@{}}
    \toprule
    \textbf{Policy}
    & \textbf{Success}
    & \makecell{\textbf{Mean steps}}
    & \makecell{\textbf{Mean pushes}}
    & \textbf{Coverage} \\
    \midrule
    \texttt{Diffusion\_Policy\_LeRobot}
    & 62.5\%
    & 223.7
    & 6.38
    & 87.4\% \\
    \texttt{Claude\_state\_controller}
    & \textbf{100.0\%}
    & \textbf{120.1}
    & \textbf{3.72}
    & \textbf{96.8\%} \\
    \texttt{Claude\_vision\_controller}
    & {97.2\%}
    & {170.4}
    & {4.76}
    & {94.2\%} \\
    \bottomrule
  \end{tabular}
\end{table}

\section{Self-Learning: Scaling to Push-A to Z}

We extend Push-T to a more challenging setting, \emph{Push-Alphabet}, replacing the single T-shaped object with 26 letter-shaped rigid bodies that exhibit diverse concavities, symmetries, and contact geometries. A single point pusher must move each object from an initial pose to a target pose using either privileged state or RGB observations. An episode is considered successful if the final target coverage reaches at least 90\% within 720 control steps. We report success rate, final coverage, mean control steps, and mean number of pushes over 25 episodes per letter. Perception details are provided in Section~\ref{sec:perception}, and full benchmark specifications are included in the appendix.


\subsection{Push-A through Push-Z}
\label{sec:selfevo}

\begin{figure}[ht]
  \centering
  \includegraphics[width=0.8\linewidth]{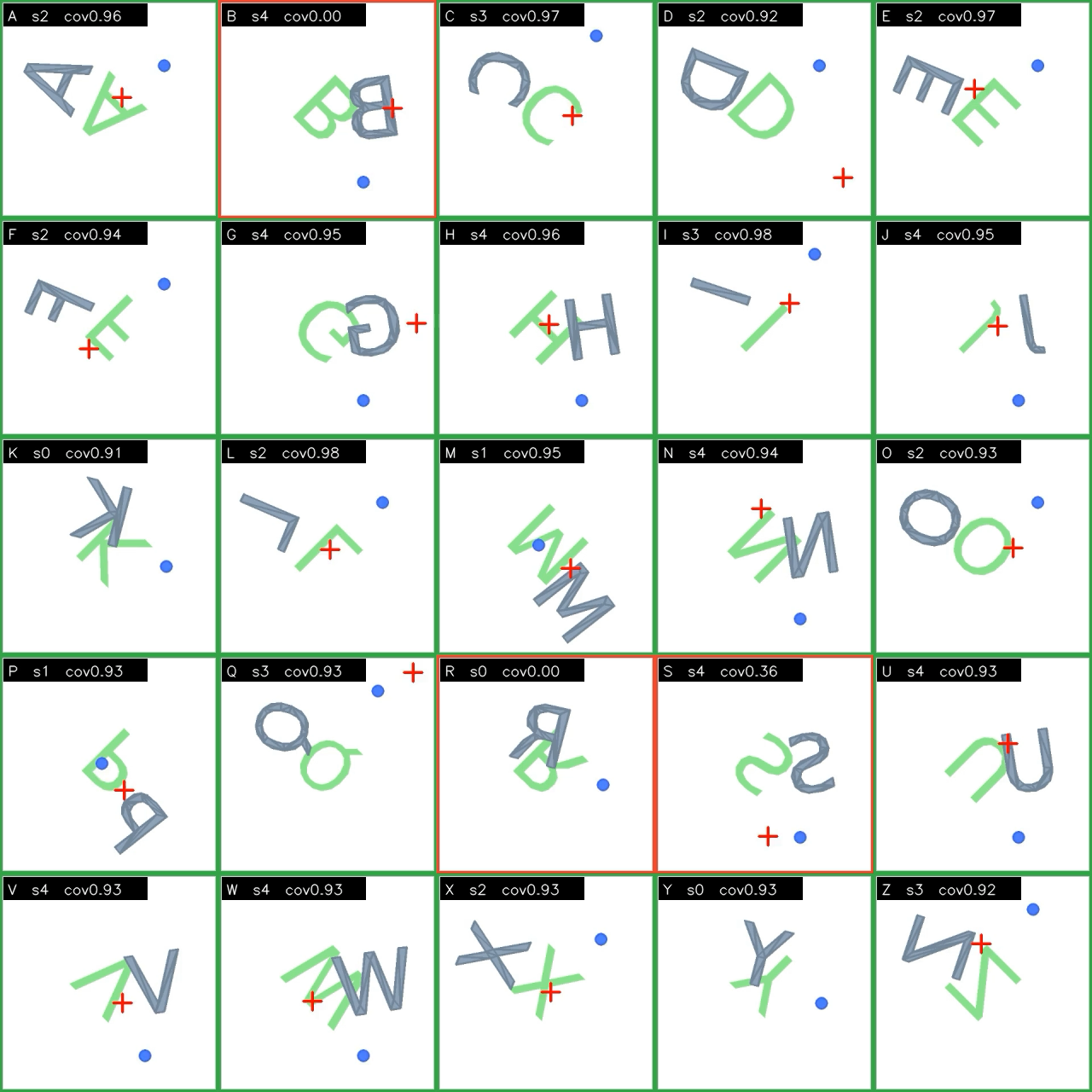}
  \caption{The agent-written Push A--Z benchmark in the 2D gym. Each tile shows the goal letter pose (green), the block (gray), and the pusher (blue dot); headers give the letter, seed, and final coverage. Red frames mark failures of the pre-MPC baseline controller (B and R at coverage 0.00, S at 0.36), which motivated the self-improvement rounds below.}
  \label{fig:az2d}
\end{figure}

Since no human demonstrations are required, we ask if agentic robotics can generalize to pushing shapes other than the letter T. We prompt the agent with Push-T to the Push A--Z benchmark, with the agent generating both the environment suite and controller iterations (Figure~\ref{fig:az2d}). 

\begin{Prompts}
    \textbf{Task Description:} Use code as policy to achieve push-alphabet. You are not allowed to use learned policy but using existing robot data is fine. Try to achieve highest success rate on all the alphabets.
\end{Prompts}

\subsection{Agent Self-Learning Process}
\begin{figure}[ht]
  \centering
  \includegraphics[width=\linewidth]{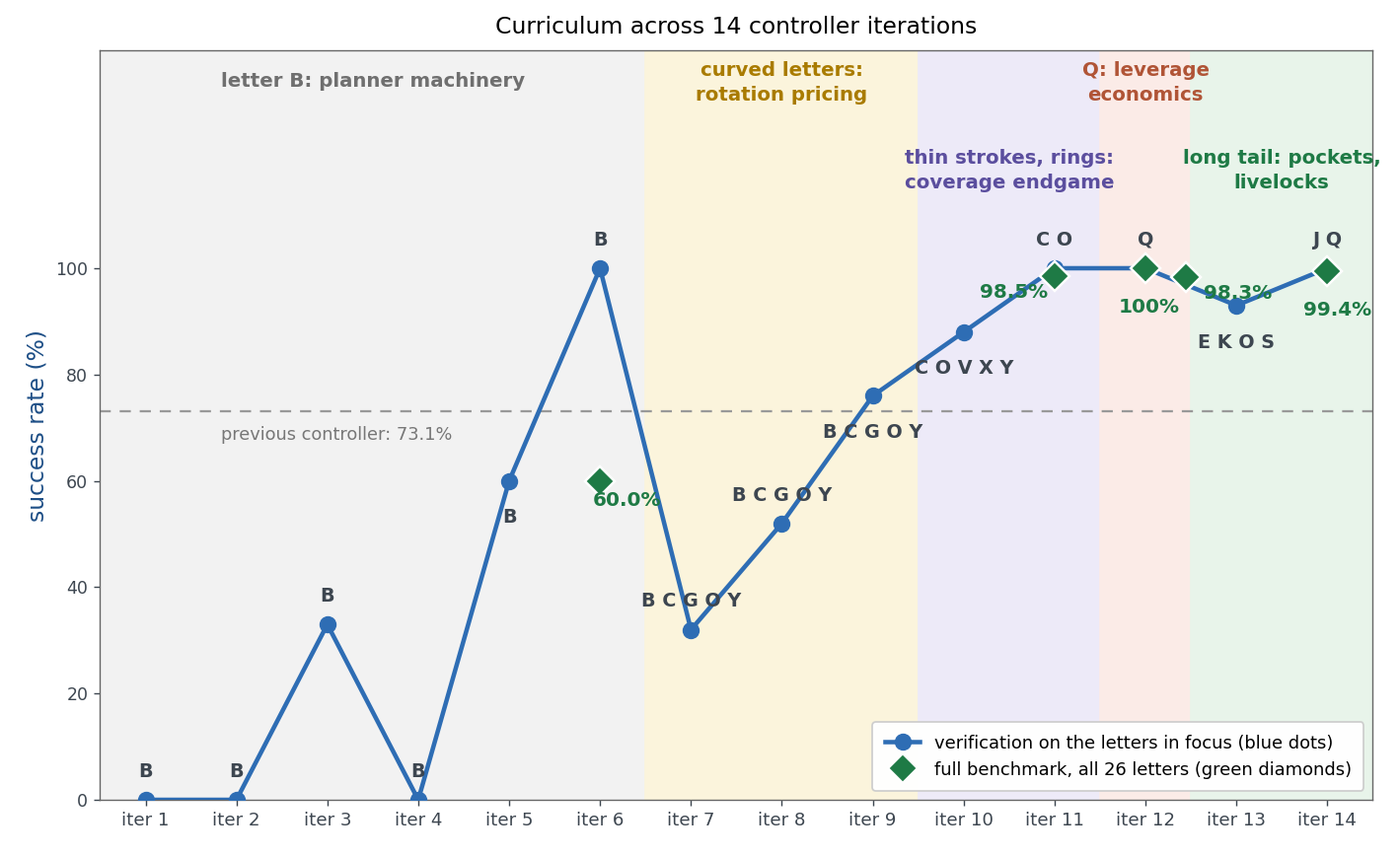}
  \caption{Self-generated curriculum across 14 controller iterations in round v3. \textbf{Blue dots} are practice tests: after each code change, the agent runs 5 episodes on the letters it is currently struggling with (those letters are printed at each dot: B; then B\,C\,G\,O\,Y; then C\,O\,V\,X\,Y; then Q; E\,K\,O\,S; J\,Q). The line dips whenever a new, harder group of letters enters the curriculum---iteration 7's crash to 32\% is not the controller getting worse but the test switching from ``just B'' to five curved letters. \textbf{Green diamonds} are exams on all 26 letters: $60.0\% \rightarrow 98.5\% \rightarrow 100\%$ (5 tries per letter), then $98.3\% \rightarrow 99.4\%$ (25 tries per letter, the strict final number).}
  \label{fig:curriculum}
\end{figure}

The agent improved its solution from push-T's \texttt{Claude\_state\_controller}
over three self-improvement iterations. In round v1, a single prompt generated an initial controller achieving a 47.1\% success rate. In round v2, autonomous self-refinement overnight elevated the success rate to 73.1\%. In round v3, the agent autonomously inspected failure logs, read the simulation source code to diagnose kinematic bugs on difficult shapes, and shifted from search-based heuristics to an MPC formulation, ultimately achieving a 99.4\% overall success rate (Figure~\ref{fig:curriculum}). Notably, operating without external web search, the agent synthesized the MPC control logic entirely from its training background.

The agent's self-generated optimization process unfolded across five distinct phases: establishing baseline planner mechanics on letter B, formulating rotation-versus-translation trade-offs for curved letters, fine-tuning corner coverage on thin-stroked characters, optimizing leverage mechanics for difficult geometries like Q, and implementing recovery rules for physical livelocks and wall-trapped states.



\noindent\textbf{Classical Baseline.}
We use Claude Code with Fable~5 to adapt the pushing method of
Zhou, Hou, and Mason~\cite{Pushing}. Under quasi-static sticking contact and
an ellipsoidal limit-surface model, the method reduces planar pushing to a
Dubins-car system, yielding bounded-curvature, time-optimal plans for a fixed
contact. We refer to this method as \emph{Dubins pushing}. These optimality properties apply to the original fixed-contact formulation; Push-Alphabet additionally requires selecting and switching contacts while keeping the pusher and object inside a bounded arena workspace. To evaluate it on
Push-Alphabet, the coding agent leaves the original algorithm unchanged and
constructs an adapter for contact selection, collision-free approach,
command conversion, and contact re-engagement. To evaluate the unchanged algorithm on Push-Alphabet, the coding agent
constructs an adapter for boundary-contact selection, collision-free
approach, conversion of contact-frame commands into pusher targets, and
re-engagement when the current contact becomes ineffective.

\noindent\textbf{Results.}
Table~\ref{tab:perception_3d} compares Dubins pushing~\cite{Pushing} with the agent-generated
MPC controllers over 650 Push-Alphabet episodes. With its initial benchmark
adapter, Dubins pushing achieves a 90.2\% success rate and 83.8\% mean
coverage. Refining only the adapter's recovery and reachability logic raises
success to 97.2\% and coverage to 90.3\%, demonstrating that the coding agent
can extend a fixed-contact theoretical controller to diverse letter
geometries without modifying its underlying algorithm. 
The agent-generated MPC controller achieves the strongest overall
performance. With privileged state input, it reaches 99.4\% success and
90.9\% mean coverage while requiring 248.7 control steps and 5.15 pushes on
average. 

Although Dubins paths are time-optimal under the original fixed-contact model, this guarantee does not extend to the complete Push-Alphabet system: contact selection, contact switching, online identification, and bounded workspace constraints are handled by the adapter and lie outside the underlying optimality result.

Compared with the refined Dubins-pushing baseline, the state-based MPC
achieves higher success and uses less than half as many pushes. For both method, the remaining failures are caused by hard reorientation episodes exceeding the finite step limit before convergence.
Dubins pushing requires more distinct contacts because each engagement includes
probing for online model identification, but its total execution remains
competitive, requiring only 18\% more control steps than MPC.

\subsection{Cross-Embodiment Transfer: Franka vs.\ UR5e}
\label{sec:3d_cross_embodiment}

\begin{figure}[ht]
  \centering
  \includegraphics[width=0.62\linewidth]{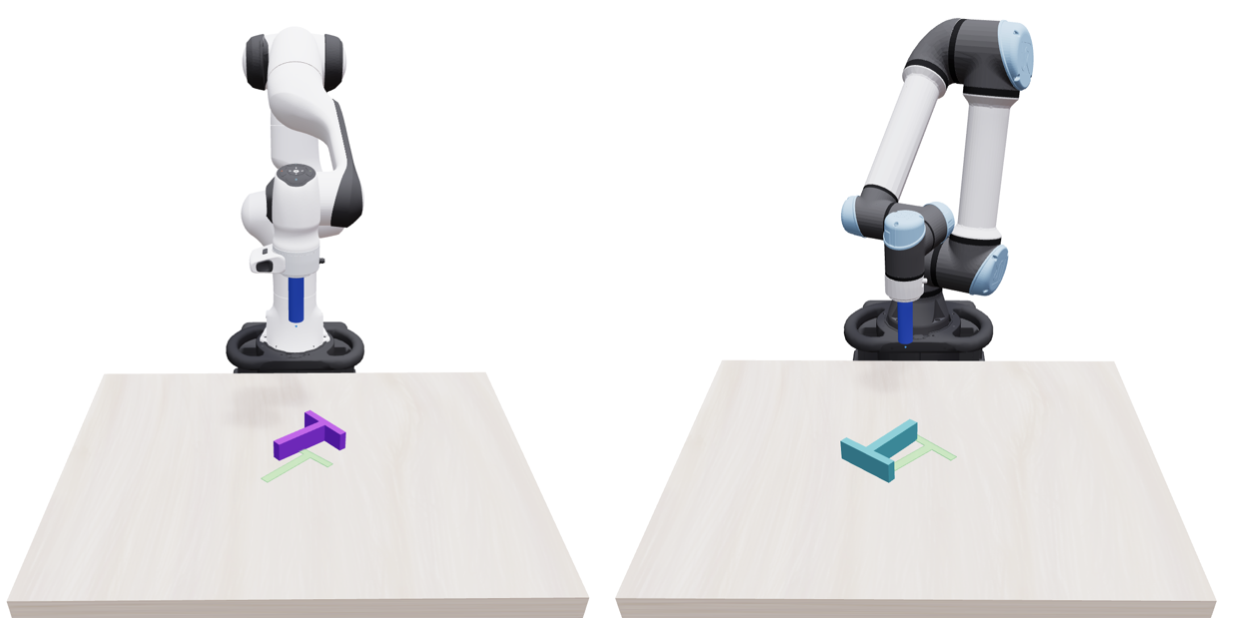}
  \caption{Franka Panda (left) and UR5e (right) pushing a letter in simulation.}
  \label{fig:arms}
\end{figure}

\begin{figure}[ht]
  \centering
  \includegraphics[width=0.78\linewidth]{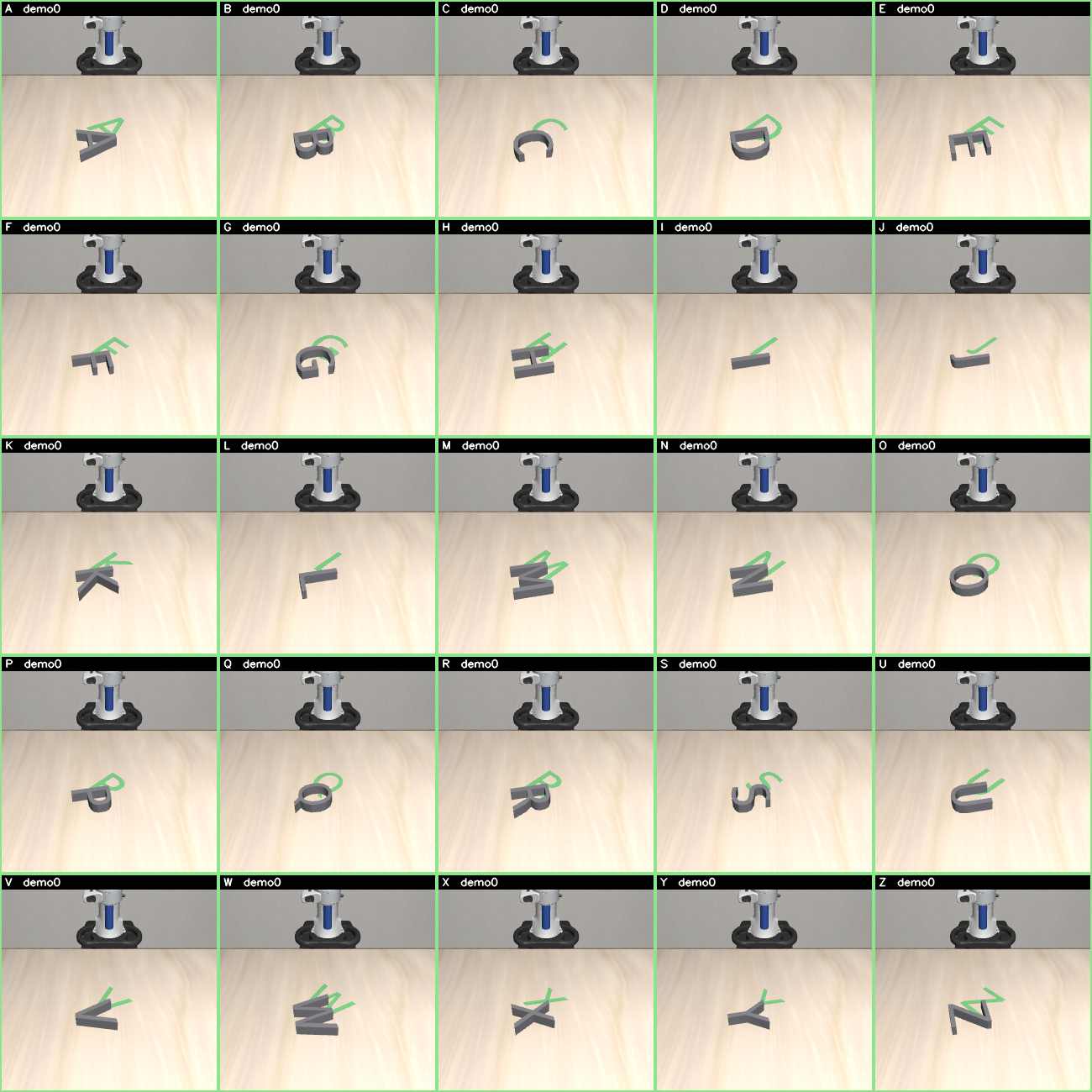}
  \caption{Push A--Z ported to simulation with a Franka arm (one tile per letter; goal letter pose in green).}
  \label{fig:az3d}
\end{figure}

We extend the evaluation of the agentic coding loop result by porting the vision-guided MPC controller into  simulation across two distinct physical embodiments: the Franka Panda and the Universal Robots UR5e arm (Figure~\ref{fig:arms}, Figure~\ref{fig:az3d}).

\begin{table*}[t]
  \caption{
  Performance comparison on Push-Alphabet.
  The classical Zhou--Hou--Mason controller, referred as Dubin Pushing~\cite{Pushing} is evaluated both
  directly and with an agent-authored adapter, while the remaining rows use
  controllers generated by Fable~5.
  Each result is evaluated over 25 episodes per letter (650 total).
  Franka and UR5e vision experiments use randomized goal poses.
  }
  \label{tab:perception_3d}
  \centering
  \small
  \setlength{\tabcolsep}{7pt}
  \renewcommand{\arraystretch}{1.08}
  \begin{tabular}{@{}llcccc@{}}
    \toprule
    \textbf{Environment} &
    \textbf{Method / Input} &
    \textbf{Mean Steps} &
    \textbf{Mean Pushes} &
    \textbf{Coverage} &
    \textbf{Success Rate} \\
    \midrule

    \multicolumn{6}{@{}l}{\textit{Classical model-based baseline}} \\
    gym
      & Dubins pushing + state
      & 291.9 & 11.00 & 83.8\% & 90.2\% \\
    \midrule
    \addlinespace[2pt]
    \multicolumn{6}{@{}l}{\textit{Classical baseline with agent-authored adapter}} \\
    gym
      & Dubins pushing + state
      & 293.8 & 12.15 & 90.3\% & 97.2\% \\

    \midrule
    \multicolumn{6}{@{}l}{\textit{Agent-generated controllers}} \\
    gym
      & MPC + state
      & \textbf{248.7} & 5.15 & \textbf{90.9\%} & \textbf{99.4\%} \\
    gym
      & MPC + vision
      & 269.7 & 5.36 & \textbf{90.9\%} & 99.2\% \\
    Franka
      & MPC + state
      & 477.0 & \textbf{4.10} & 90.3\% & 99.2\% \\
    Franka
      & MPC + vision
      & 668.1 & 5.78 & 83.4\% & 89.8\% \\
    UR5e
      & MPC + vision
      & 545.8 & 5.06 & 90.0\% & 98.5\% \\

    \bottomrule
  \end{tabular}
\end{table*}

\textbf{Result:} 
The vision MPC achieves a 98.5\% success
rate on the simulated UR5e and an 89.8\% success rate on the simulated Franka
Panda, requiring $5.1$ and $5.8$ mean number of pushes, respectively. These results
provide preliminary evidence that the generated controller is portable across
distinct simulated embodiments without modifying its task-level contact
reasoning. 


\section{Model Cost}
Across the code-policy, MPC, vision model, and agent-facing benchmark components, the coding agent generated about 9.5 million output tokens. Because much of the workflow reused cached context, the total API cost was approximately \$1,500--\$2,000. The agentic development process required approximately 221 hours in Claude ``auto mode'' under human interactions. 

\section{Limitation and Discussion}
One limitation concerns how the coding agent interprets the simulation environment. General-purpose coding agents typically treat the provided sandbox as a complete and authoritative specification of the task. In robotics, however, the simulator is only an approximation of the physical system. The agent may exploit simulator-specific APIs, state variables, rendering conventions, contact behavior, or numerical regularities that are unavailable or inaccurate on a physical robot. Although the generated controller performs parameter adaptation within simulation and transfers across simulated embodiments, this does not guarantee robustness to real-world perception errors, calibration drift, latency, compliance, friction variation, or unmodeled contacts. Future agentic robotics systems should therefore treat simulation as an uncertain hypothesis rather than a fully specified environment.

Claude is computationally expensive and sensitive to the prompt. Each refinement stage can consume more than million language-model tokens across code generation, simulator inspection, execution analysis, debugging, and repeated controller revision. Running the complete pipeline with a frontier coding model such as Fable 5 may therefore incur substantial inference cost. An important direction is to determine which stages require a frontier model and which can be delegated to smaller or less expensive models, for example by using a strong model only for high-level diagnosis and architecture changes while using cheaper models for localized code edits, log analysis, and parameter tuning.
\bibliographystyle{unsrt}
\bibliography{references}

\newpage
\section{Appendix}

\subsection{The Push-Alphabet Benchmark}
\label{sec:pushalphabet-setup}

Push-Alphabet generalizes Push-T from a single canonical block to the 26
letter shapes A--Z, while keeping the physics, the pusher, and the success
criterion identical to the original \texttt{gym\_pusht} environment. Each
letter environment is a subclass of \texttt{PushTEnv}, so a controller that
runs on Push-T runs on Push-Alphabet without modification; only the block
geometry changes.

\paragraph{Letter geometry.}
Each block is the filled outline of one letter in the DejaVu Sans font,
extracted as a polygon at runtime, scaled to a height of 140\,px, and
recentered on its area centroid so the body frame origin coincides with the
center of mass. The outline is simplified with a 2\,px tolerance and
convex-decomposed into triangles for the physics engine; interior holes
(e.g., A, O) are preserved. The resulting family spans near-convex strokes
(I, L), deep concavities (C, G, U), thin open strokes (V, W, Y), and
holed shapes---a substantially harder range of contact geometries than the
single T block. The block has unit mass, friction coefficient 1, and its
moment of inertia follows from the uniform-density polygon.

\paragraph{Physics and control.}
The arena is a $512 \times 512$ workspace bounded by walls. The pusher is a
disc of radius 15\,px driven by a PD controller
($k_p{=}100$, $k_v{=}20$) toward an absolute target position; the agent
acts at 10\,Hz (10 physics substeps of 10\,ms per control step), and the
action space is the target position
$a_t \in [0, 512]^2$. The pusher therefore lags its commanded target,
and its realized speed drops by roughly an order of magnitude when pushing
against the block---controllers must cope with this actuation gap rather
than command idealized velocities.

\paragraph{Episodes.}
Each episode is seeded: the block pose (position in the central region,
orientation uniform in $[-\pi, \pi]$) and the initial pusher position are
drawn from the environment's per-seed RNG, so all methods see identical
initial conditions. The target pose is the canonical \texttt{gym\_pusht}
goal $(256, 256, \pi/4)$, retained so the pretrained diffusion baseline
remains meaningful; a randomized-goal mode is also available. An episode
succeeds when the \emph{coverage}---the area of the intersection between
the current and target letter footprints, normalized by the target
footprint area---reaches $0.9$, and terminates on success or after 720
control steps.

\paragraph{Observations and visibility.}
In the \emph{state} setting the observation is
$o_t = \left[p^a_x,\, p^a_y,\, p^\ell_x,\, p^\ell_y,\, \theta^\ell\right]$,
the pusher position and letter pose; the \emph{vision} setting provides RGB
frames from which these quantities must be estimated. Independently of the
observation stream, a per-episode task specification is passed to the
controller before the first action, redacted by a visibility level:
\emph{full} (letter polygon, pusher radius, inertia, friction, PD gains),
\emph{shape} (polygon and radius only), or \emph{blind} (no geometry).
Controllers interact with the benchmark only through this specification and
the observation--action loop; they never access the simulator state
directly.

\paragraph{Metrics and protocol.}
We report \emph{success rate}, \emph{mean control steps}, and \emph{mean
number of pushes}. Pushes are counted on the simulator side by a hysteresis
(Schmitt-trigger) detector on the ground-truth pusher--block surface gap
(contact entered below $0.5$\,px, broken above $3$\,px for at least two
consecutive steps), making the push count independent of any
controller-internal bookkeeping. Unless stated otherwise, evaluations run
25 episodes per letter with seeds $0$--$24$ ($650$ episodes total). 

\end{document}